\RequirePackage{fix-cm}
\documentclass[smallextended]{svjour3}       
\smartqed  
\usepackage{graphicx}
\usepackage{url}
\usepackage{subfig}
\usepackage{gensymb}
\begin{document}

\title{Facial Landmark Detection for Manga Images}


\author{Marco Stricker \and Olivier Augereau \and Koichi Kise \and Motoi Iwata}


\institute{Marco Stricker \at
              University of Kaiserslautern \\
           \and
           Olivier Augereau, Koichi Kise, and Motoi Iwata \at
              Osaka Prefecture University
}

\date{Received: date / Accepted: date}

\maketitle

\begin{abstract}
The topic of facial landmark detection has been widely covered for pictures of human faces, but it is still a challenge for drawings.
Indeed, the proportions and symmetry of standard human faces are not always used for comics or mangas. 
The personal style of the author, the limitation of colors, etc. makes the landmark detection on faces in drawings a difficult task. 
Detecting the landmarks on manga images will be useful to provide new services for easily editing the character faces, estimating the character emotions, or generating automatically some animations such as lip or eye movements.

This paper contains two main contributions: 1) a new landmark annotation model for manga faces, and 2) a deep learning approach to detect these landmarks. We use the "Deep Alignment Network", a multi stage architecture where the first stage makes an initial estimation which gets refined in further stages.
The first results show that the proposed method succeed to accurately find the landmarks in more than 80\% of the cases. 
\end{abstract}

\section{Introduction}

Drawings are used to represent real or imaginary situations, persons, objects, ideas, etc. 
For the case of painting, some art uses realistic styles such as realism or hyperrealism, but many art styles manipulate the shapes, colors and textures to give special emotions or expressions such as expressionism, cubism, surrealism, etc. 
As compared to real word pictures, drawings contain a much wider variety of content and is more complex to analyze~\cite{inoue2018cross}.      

In a similar way, mangas (Japanese comics) have a large variety of styles, content, and genres such as comedy, historical drama, science fiction, fantasy, etc. 
Detecting the characters in manga images is not a trivial problem because the protagonists of the stories are quite various, such as: humans, animals, monsters, etc. And even when the characters are humans, their face can have large deformations~\cite{chu2017manga}.  


The manga market in Japan is very large. In 2016, The All Japan Magazine and Book Publisher's and Editor's Association (AJPEA) reported\footnote{\url{http://www.ajpea.or.jp/information/20170224/index.html}} that the sales of the manga in Japan reached approximately 3.9 billion USD. In this report we can also see that the the digital market almost doubled between 2014 and 2016.  

The analysis of comics and mangas images recently sparked the computer vision and document analysis communities interest~\cite{augereau2018survey}.
The digital version of manga can be used by the researchers to propose new algorithms to provide services such as dynamic visualization of manga~\cite{augereau2016comic}, adding colors~\cite{kataoka2017automatic}, generating animations ~\cite{gupta2018imagine}, creating new kinds of recommender systems~\cite{daiku2017comic}, etc.

In this paper we will start by giving a brief overview of the current state of the art in landmark detection and the work which has been done in the field of manga faces in Section~\ref{sec:soa}. 
Section~\ref{sec:dataset} will be dedicated to the introduction of our new landmark annotation model for manga faces, and the dataset we created. 
Following this, Section~\ref{sec:method} will give a brief overview about the Deep Alignment Network~\cite{kowalski2017deep} which we have used for detecting facial landmarks on manga images. 
Section~\ref{sec:experiment} details all the experiments and how we evaluate our experiments. The results are described in Section~\ref{sec:results}. We will end our paper with a conclusion in Section~\ref{sec:conclusion}.

\section{State of the art}\label{sec:soa}

\subsection{Landmark detection}

Facial landmarks detection, or face alignment consists in localizing several parts of the face in an image such as the eyebrows, eyes, mouth, nose, chin, etc. 
This task has been heavily investigated for human faces, and is still a very active research domain as it can be seen in recent publications~\cite{kowalski2017deep,bulat2017far,lv2017deep,wu2017simultaneous}.
There are different approaches for detecting facial landmarks. 
Before the success of deep learning, researchers used shape indexed features~\cite{xiong2013supervised,ren2014face}. This approach focused on extracting image local features.
The latest publications are focusing on deep learning approaches, using mostly CNNs (Convolutional Neural Networks). 
The emergence of large databases, which are crucial for deep learning approaches, plays an important role in face alignment~\cite{sagonas2016300,sagonas2013semi}.
Instead of using local patches, the Deep Alignment Network method proposed by Kowalski et al.~\cite{kowalski2017deep} is based on the entire image, and consists of multiple stages. Each stage takes the output of the earlier stage as an input, and tries to refine it.
Another approach proposed by Lv et al.~\cite{lv2017deep} is improving the initial estimation of landmarks using a deep regression architecture.

A problem in face alignment is facial occlusions, which has being tackled by Wu et al.~\cite{wu2017simultaneous}. They combined landmark detection, pose, and deformation estimation in one approach in order to make the landmark detection robust to occlusions. 
Aside from the many approaches for human faces, Rashid et al. ~\cite{rashid2017interspecies} proposed a method for animal faces. In this method the animal face is being warped to resemble to a human face, so that normal human face detectors may work on it.

\subsection{Landmark models}
Several different models to label the facial landmarks have been proposed. 
One of the simplest models utilizes only five landmarks~\cite{zhang2016joint}: two in the center of the eyes, one for the nose and two for the corners of the mouth. 
One of the most complex models is based on 194 control points~\cite{kasinski2008put,le2012interactive} corresponding to the face, nose, eyes, eyebrows, and mouth outlines. 

However the iBUG~\cite{sagonas2016300} model based on 68 landmarks is being used by the 300W~\cite{sagonas2016300} and Menpo~\cite{zafeiriou2017menpo} challenges seems to be the current state of the art. This model is also used by other libraries such as OpenFace~\cite{amos2016openface}. 

\subsection{Manga image analysis} 

For the manga and comics image analysis, some related research has been done for generic object recognition in drawing~\cite{inoue2018cross} and more specifically the character face detection~\cite{chu2017manga,jha2018towards}. 

The landmark detection for cartoon images has been introduced by Jha et al.~\cite{jha2018towards}. 
The dataset consists of caricature images with different painting style and is available online. The landmark annotations of 750 face images have been made available by the authors. 
The authors defined 15 landmarks for one face: six for the eyes, four for the eyebrows, one for the nose and four for the mouth. Surprisingly, the authors achieved better performance for the landmark detection by using only real face images in the training set. Using only caricature images or mixing caricature images and real face images degraded the performances of their system.

\section{New landmark model}\label{sec:dataset}
As drawn faces have more variations than real human faces, we propose to use 60 landmarks, including landmarks for the chin contour.  
In this section we will describe our dataset and our landmark model.

\subsection{Data Sources}
Availability of manga images which can be used for research and shared publicly are scarce because of copyright issues. 
However Matsui et al.~\cite{matsui2017sketch} released "Manga109", a dataset of 109 manga volumes for research purposes. 

Chu et al.~\cite{chu2017manga} used this dataset to train a CNN for detecting the face bounding boxes on manga images. 
To achieve this, they manually labeled the faces in 66 volumes. They have used 50 pages per title and published a subset of this dataset, which can be freely downloaded~\cite{MangaFace}. 
This subset contains face bounding boxes from 24 different volumes. We have used these face bounding box information in order to extract the faces from the manga pages.

\subsection{Image Selection}
From Chu et al. dataset we could extract 5505 face images. However we decided to remove some of these images. 
First of all, a face may appear in different poses. A major difference exists between a frontal face which shows all facial features and a profile view of the face which only shows half of the features. 
Therefore, in the current state of the art (iBUG model~\cite{zafeiriou2017menpo}), two different annotation models exist to describe frontal faces and profile faces. 
We decided to focus on frontal faces and therefore removed manga faces with a profile view.

After that, we removed all images whose resolution was too small to correctly annotate the landmarks. 
In an image, the facial features should be clearly distinguishable by a human. 
If the width or height of an image is smaller than 80 pixels, the image has been removed. 

In contrast to human faces which follow regularities, the creativity of a manga authors knows no bounds and therefore faces may have inhuman features and can look like something we would not consider human. 
The face image need to have exactly two eyes, one nose, one mouth and one chin visible, otherwise it was discarded.  

Lastly we removed faces where not both eyes were visible due to glasses or their hair.
Our dataset is finally consisting of 1446 images of faces.

\subsection{Landmarks Model}\label{sec:landmarkModel}
\begin{figure}
	\centering	
	\includegraphics[scale=0.3]{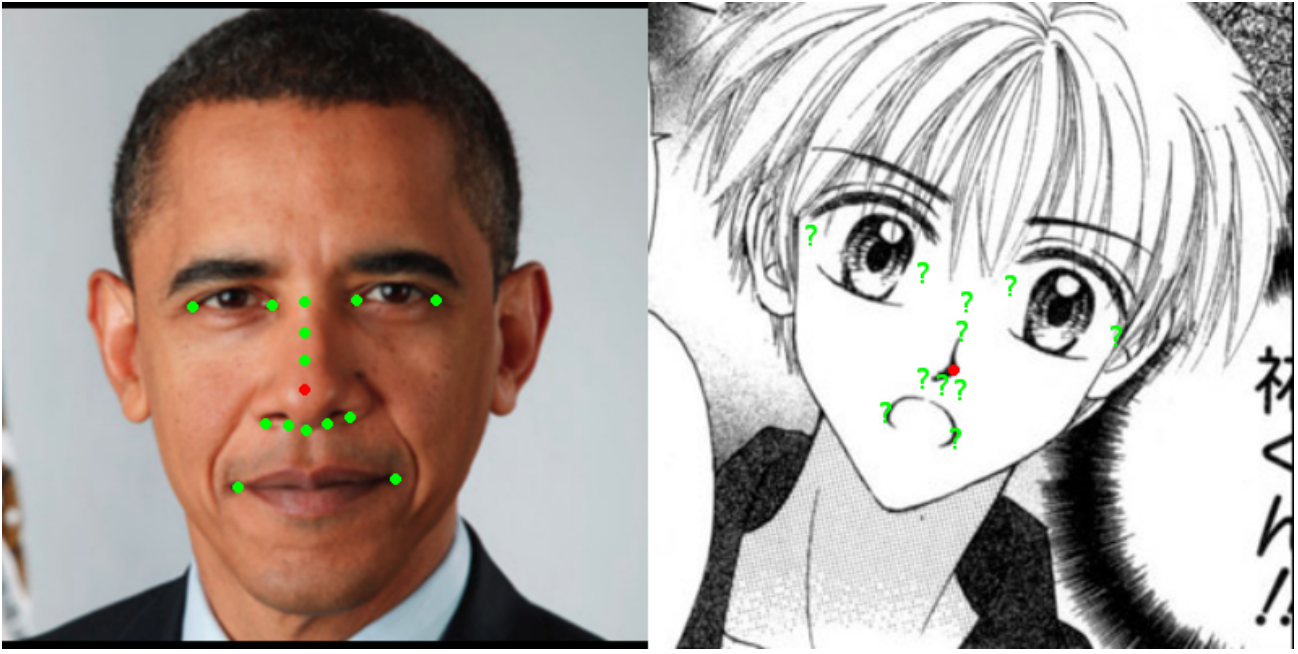}
	\caption{Differences between the landmarks of a human face and a manga face. The corners of the eyes and mouth are not drawn. The bottom part of the nose is also missing. Manga face from ``PrismHeart'' by Asatsuki Mai}
	\label{fig:landmarkDifferences}
\end{figure}

We decided to employ a similar model as the iBUG model~\cite{sagonas2013300} with 68 landmarks. 
Unfortunately a one to one transition is not possible, since not all features that can be found on a human face can be found on a manga face so we adapted the iBUG model for the manga face images by removing some landmarks and adding some other ones. 
Figure~\ref{fig:landmarkDifferences} highlights some typical landmarks that usually appear in a human picture but not in a manga drawing.
In manga, the nose of characters is usually simplified in a line (so it is difficult to define an outer part and inner part). The same goes for the mouth, where the lips are usually not drawn but simplified in a line. 
The corners of the eyes and the mouth are not always clearly drawn in manga images. 

\begin{figure}
	\centering	
	\includegraphics[scale=0.3]{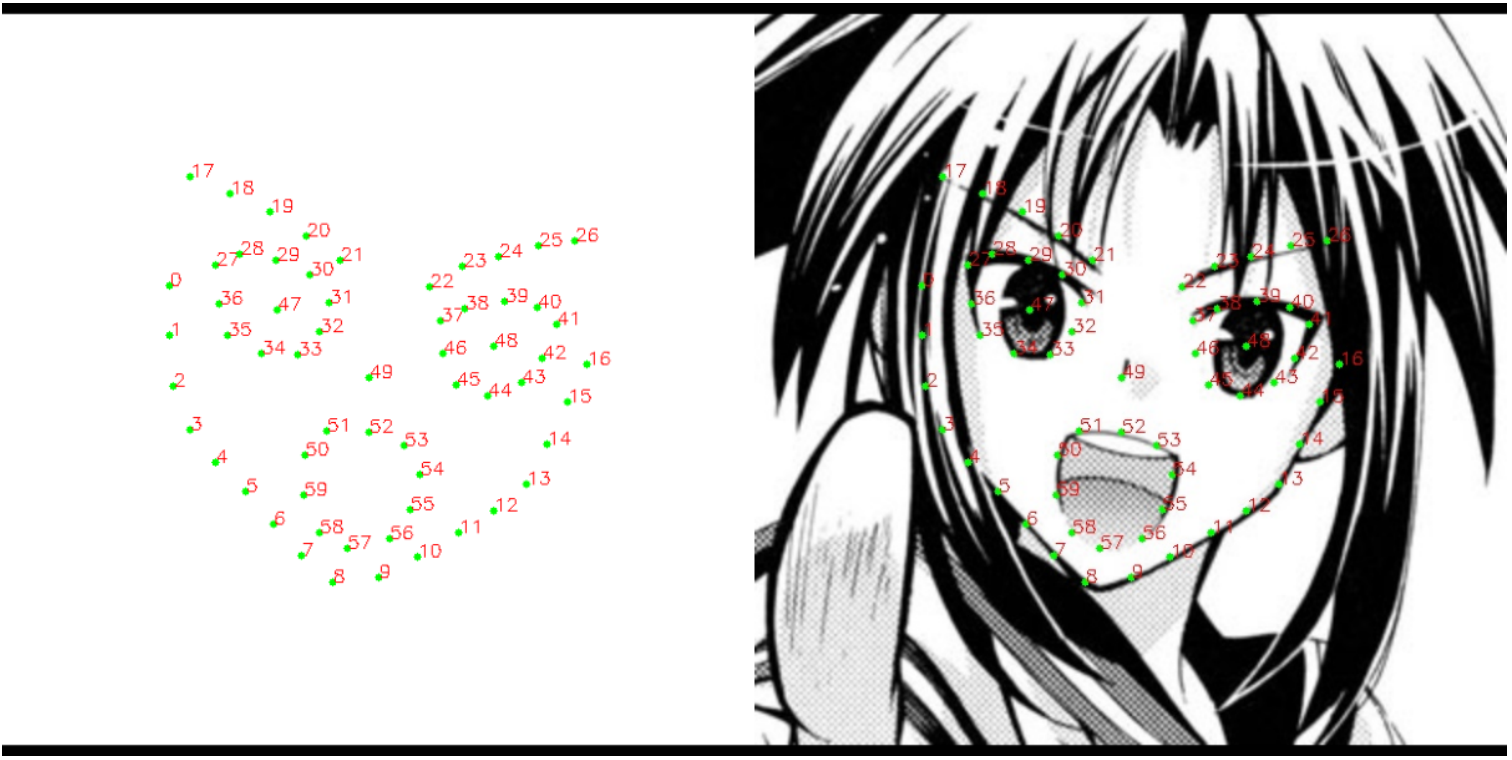}
	\caption{An example showing how the final landmark model looks like. Manga face taken from ``MeteoSanStrikeDesu'' by Takuji}
	\label{fig:landmarkExample}
\end{figure}

In the iBUG model, the landmarks for the eyes are spread equidistantly on both eyelids starting from the intersection of the eyelids. 
On manga faces, this intersection is rarely drawn. 
Since many characters have round shaped eyes, we applied 10 landmarks on each eye which are spread equidistantly. 
This results in an accurate representation of the contour of the eyes, as it can be seen in Fig. \ref{fig:landmarkExample}. 
We decided to add one landmark to define the pupil of each eye of the character.

We applied the same technique for the mouth and eyebrow of a character. Since they mostly consist of one line in manga images, we just use 10 landmarks to define the line of the lip and 5 for each eyebrow. 

Noses in manga faces are mostly simplified compared to human features. 
The nostrils are not drawn and the nose dorsum line is sometimes only implied. 
Therefore, we also decided to simplify our landmarks for the nose compared to the iBUG model and use only one landmark to define the tip of the nose, as it can be seen in Fig.~\ref{fig:landmarkExample}. 

Lastly we followed the landmarks positioning proposed by the iBUG model for the chin contour. 
To sum up, our landmark model consists of the following 60 landmarks:
\begin{itemize}
	\item 5 landmarks for each eyebrow
    \item 10 landmarks for each eye
    \item 1 landmark for each pupil
    \item 1 landmark for the nose
    \item 10 landmarks for the mouth
    \item 17 landmarks for the chin contour
\end{itemize}

\subsection{Labeling Process}

Each of the 1446 images of our dataset has been labeled by at least one participant. 
To control the quality of the labeling, 656 images were labeled twice.
If the distance between two landmarks was greater than 2 pixels, then these landmarks were manually corrected and compared again. 
With this procedure, wrong landmarks could be found and corrected. 
Then the two labels are merged by computing the spatial average for each landmark. 

The participants were asked not to put labels if they were visually not present.
If the chin contour, the mouth contour and two eyes are labeled, the missing parts have been added automatically: (1) the nose is calculated as the mass center of the two eyes and the mouth, (2) the pupil is computed as being the mass center of two landmarks of the eye. 
(3) If one eyebrow is missing, then a mapping matrix is computed between the eye landmarks, which describes how to map the landmarks from one eye to the other. After that the existing eyebrow is used with the mapping matrix to calculate the missing eyebrow. 
(4) If both eyebrows are missing, the landmarks which define the upper eyelid are copied, scaled and translated above the eyes.

\section{Proposed method}\label{sec:method}

For the landmark detection we use an adapted version of the Deep Alignment Network~\cite{kowalski2017deep}(DAN). 
The DAN was published in 2017 and achieves state of the art results for human face alignment in pictures. 
We adapted the DAN to change the amount of landmarks. As described in section \ref{sec:landmarkModel} we use a landmark model with 60 landmarks instead of the 68 iBUG model for which the DAN was initially build. 
We are now going to briefly summarize the network.

The DAN is a multi-stage network and uses an approach similar to Cascade Shape Regression (CSR)~\cite{yang2015facial}. 
It initializes a first face shape $S_{0}$ which gets improved in the next stages of the network. In this model, each stage performs feature extraction and regression. A difference between DAN and most CSR methods is, that CSR methods perform feature extraction on patches around the landmarks. DAN uses a landmark heat map as an input in order to extract features from the whole image. 

The input of the first stage is the input image. This stage calculates a landmark heatmap $H_{t}$ and a feature image $F_{t}$. After that $H_{t}$, $F_{t}$ and the input image are the input for the second stage. This stage also calculates $H_{t}$ and $F_{t}$ which is then forwarded to the next stage. This procedure is repeated for the number of stages. 

Each stage calculates the landmark locations and the input for the next stage. The landmark location is handled by a feed-forward neural network, while connection layers are responsible for the input for the next stage. 

During the training process, DAN starts training the first stage until the validation errors stops improving. After that, the next stage is included into training until the validation error stops improving again and so forth. 


\section{Experiment}\label{sec:experiment}
With the DAN and the above described dataset we performed multiple experiments. 
Since the DAN can be trained for multiple stages we have trained it for one and for two stages. All experiments have been trained for 150 epochs.

\subsection{Split}

The dataset is divided into training, test, and validation set by a random split. 
With a random split, the faces are distributed randomly into training-/test- and validation-set. Therefore each set will likely contain faces from every title. We divided the images into 80\% training set, 10\% validation set and 10\% test set.


\subsection{Augmentation}
In order to avoid overfitting and to increase the accuracy of the model, we applied data augmentation and increased our training set size by a factor of five. An image got randomly augmented by translation, rotation and scaling. The number by how much an operation was applied is a random number, which follows a Gaussian probability density with different standard deviations as described in equation~\ref{eq:gaussian}, where $\mu$ is the mean and $\sigma$ is the standard deviation. We applied five random transformations for each image.

\begin{equation}
p(x)=\frac{1}{\sqrt{2\pi\sigma^{2}}}e^{-\frac{(x-\mu)^{2}}{2\sigma^{2}}}
\label{eq:gaussian}
\end{equation}

For the different transformations we have used following $\mu$'s and $\sigma$'s:
\begin{itemize}
\item rotation angle in degrees: $\mu=0$ and $\sigma=20$
\item scaling factor: $\mu=1$ and $\sigma=0.1$
\item translation factors $t_{x}$ and $t_{y}$: $\mu=0$ and $\sigma=0.1$
\end{itemize}

Furthermore the translation factors $t_{x}$ and $t_{y}$ are multiplied by the width and height of the scaled mean shape. The mean shape is calculated by averaging over all landmarks in the training set. Scaling it into a sub-rectangle of the image defined by certain margins results in the scaled mean shape. 

Therefore our translations are within rotations of $-20\degree$ and $20\degree$, the scaling is within 90\% and 110\% and the translations are independent from the image size.
To sum it up, we have run a total of 4 experiments. The combinations are shown in table~\ref{table:experiments}.

\begin{table}[h]
  \centering
  \caption{All combinations of experiments we have run}
  \begin{tabular}{ | c | c | }
    \hline
    Data Augmentation   & Stages  \\ \hline
    Yes & 1 \\ \hline
    Yes & 2 \\ \hline
    No &  1\\ \hline
    No &  2\\ \hline
  \end{tabular}
  \label{table:experiments}
\end{table}

\subsection{Evaluation}\label{sec:eval}
The evaluation is done by calculating the distance between a ground truth landmark and the corresponding predicted landmark normalized by a certain value $v$. 
For human faces, the normalization factor $v$ is often chosen to be the distance between the two eye mass centers. However we decided to introduce a new distance for manga faces. 

\begin{figure}
    \centering
    \includegraphics[width=.5\textwidth]{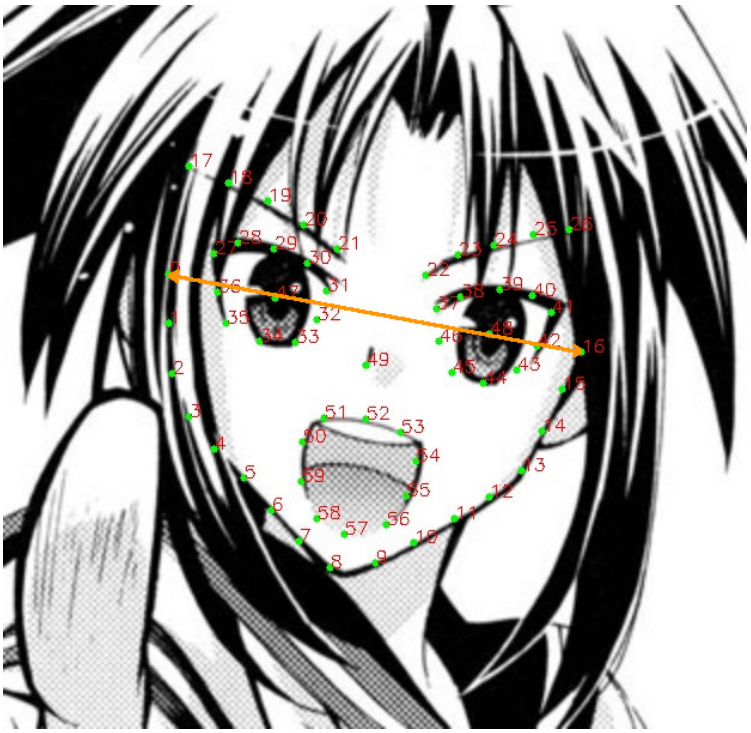}
    \caption{Illustration of the chin normalized Distance. Manga face taken from ``MeteoSanStrikeDesu'' by Takuji}
    \label{fig:normalization}
\end{figure}

We decided to use the \textit{chin normalized distance} instead of the distance between the two eye mass centers. The distance between the two eye mass centers is used, because due to genetic reasons human eyes mostly follow the same face proportions. However this is not true for manga faces. Manga eyes proportions and sizes vary a lot, so that the reader can easily distinguish different characters. In order to avoid such cases, which might influence the performance, we decided to use the distance between the first and the last landmark of the chin contour "\textit{chin normalized distance}" as a normalization factor. This distance is also used as a loss function for training.
The \textit{chin normalized distance} can be seen in Fig.~\ref{fig:normalization}.

For one image, if the average distance (or in other words mean error) over all landmarks is greater than a threshold, then these predicted landmarks are considered a failure.
Furthermore we calculated the area under the cumulative distribution curve calculated up to a threshold $\alpha$ and then divided by that threshold (denoted $A_{\alpha}$).

In detail we calculate the average distance between ground truth landmarks and predictions on a single face with:

\begin{equation}
    \bar{D}=\frac{1}{N}\sum_{n=1}^{N}d(\hat{y}_{n},y_{n}),
\end{equation}

where $N$ defines the total number of landmarks, $\hat{y}_{n}$ is the prediction coordinate, $y$ is the ground truth coordinate and $d(x,y)$ the Euclidean distance. 
The \textit{chin normalized distance} $D_{chin}$ is the euclidean distance between the first chin landmark and the last chin landmark as it can be seen in Fig.~\ref{fig:normalization}. Therefore the normalized average distance $S$ per face is:

\begin{equation}
    S = \frac{\bar{D}}{D_{chin}}
\end{equation}

Lastly we consider an image a failure if the $S$ is bigger than the threshold of $0.0333$. During the manual labeling phase, we have labeled some images twice. The largest distance between two sets of landmarks for one image that we have encountered was $0.0333$. 
Therefore humans still consider this distance as the same point. This means, that if our automatic prediction is within this range we can say that this was a success, or at least within human capabilities. The failure rate defines the percentage of images which are considered a failure.

\section{Results}\label{sec:results}

\begin{figure}
    \centering
    \subfloat[][Augmented with random split]{\includegraphics[width=.5\textwidth]{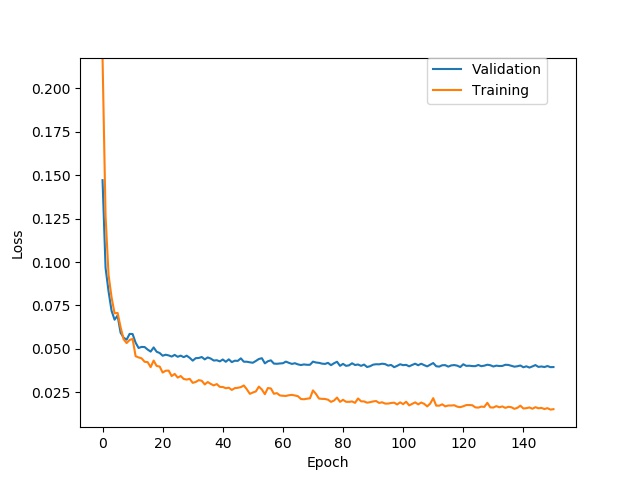}\label{subfig:oneAugRSError}}
     \subfloat[][No augmentation with random split]{\includegraphics[width=.5\textwidth]{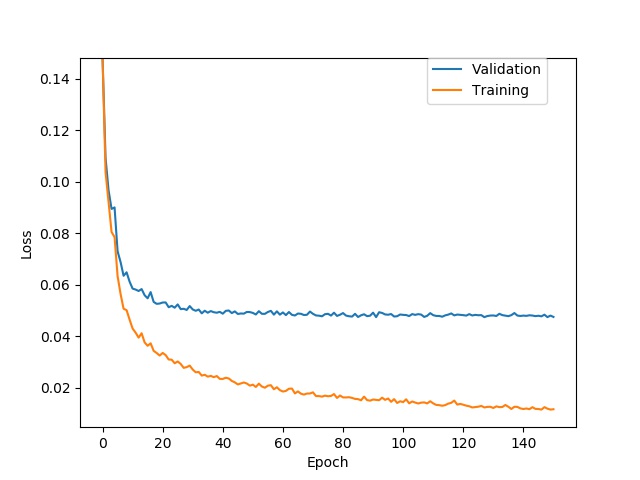}\label{subfig:oneNaRSError}}
    \caption{Loss curves for a one stage network}
    \label{fig:lossCurves1}
\end{figure}

\begin{figure}
    \centering
    \subfloat[][Augmented with random split]{\includegraphics[width=.5\textwidth]{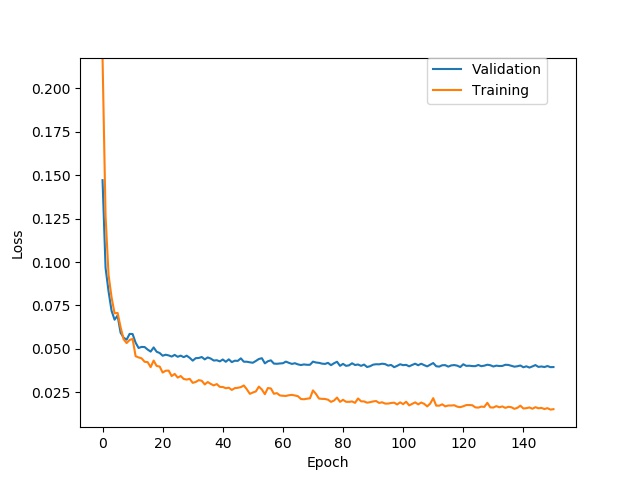}\label{subfig:TwoAugRSError}}
     \subfloat[][No augmentation with random split]{\includegraphics[width=.5\textwidth]{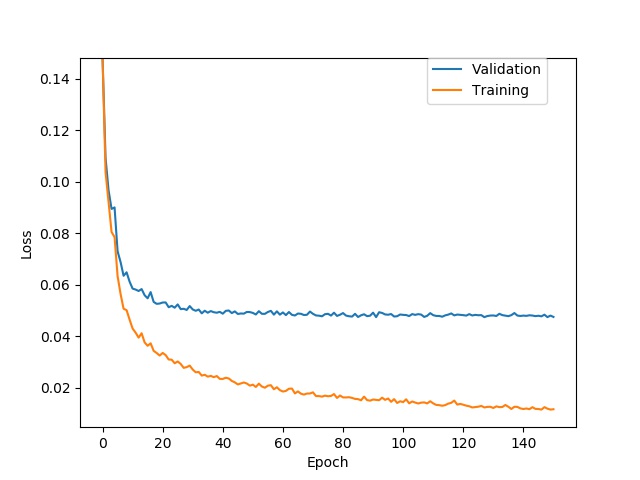}\label{subfig:TwoNaRSError}}
    \caption{Loss curves for a two stage network}
    \label{fig:lossCurves2}
\end{figure}

First of all we will investigate the training loss curves. For that see figure~\ref{fig:lossCurves1} for a one stage network and figure~\ref{fig:lossCurves2} for a two stage network.
As we can see the overall loss on the two stage DAN is lower compared to the one stage DAN. But in both stages the validation loss is significantly higher than the training loss. This means that the network does not generalize well and performs worse on unseen samples. The figures also show that this gap gets smaller if data augmentation is applied. Therefore we assume that our network can still perform better if it has access to more data.

Table~\ref{table:results} shows the mean error, the $A_{0.0333}$ and the failure rate for all experiments we have run. As it can be seen, the model trained for two stages with data augmentation and random split is the best performing model. In our case data augmentation and two stages also improved the performance. Lastly we also experimented with a three stage model, however it did not improve the performance anymore.

\begin{table}[h]
	\centering
	\begin{tabular}{ | c | c | c | c | c |}
		
		\hline		
		Stages & Augmentation & Mean Error & $A_{0.0333}$ & Failure Rate (\%)\\ \hline
		1 & Yes & 0.03933 & 0.10338 & 48.28 \\ \hline
		1 & No & 0.04355 & 0.08964 & 52.41 \\ \hline
		2 & Yes & 0.02935 & 0.24295 & \textbf{19.31} \\ \hline
		2 & No & 0.03467 & 0.16357 & 37.93 \\ \hline
		
	\end{tabular}
	\caption{Results of the experiments. Highlighted is the best failure rate}
    \label{table:results}
\end{table}

\begin{figure}
\centering
\includegraphics[width=.4\textwidth]{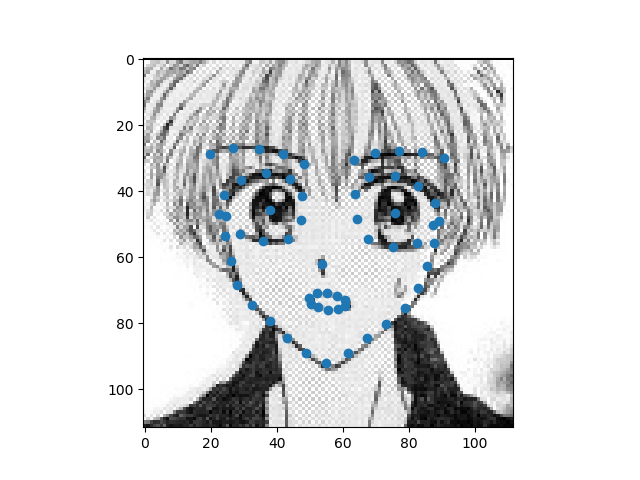}
\includegraphics[width=.4\textwidth]{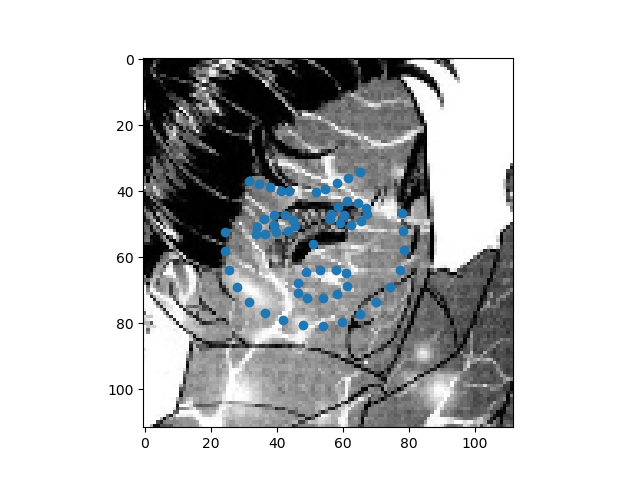}\\
\includegraphics[width=.4\textwidth]{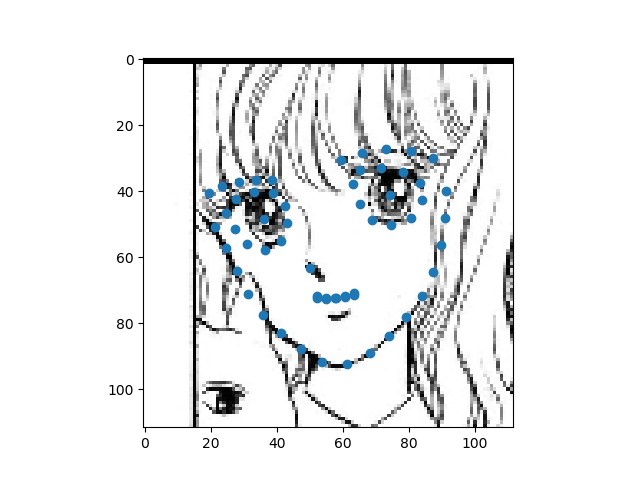}
\includegraphics[width=.4\textwidth]{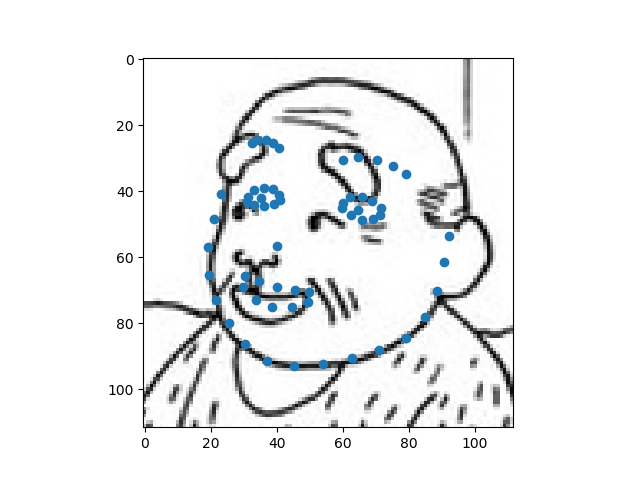}\\
\includegraphics[width=.4\textwidth]{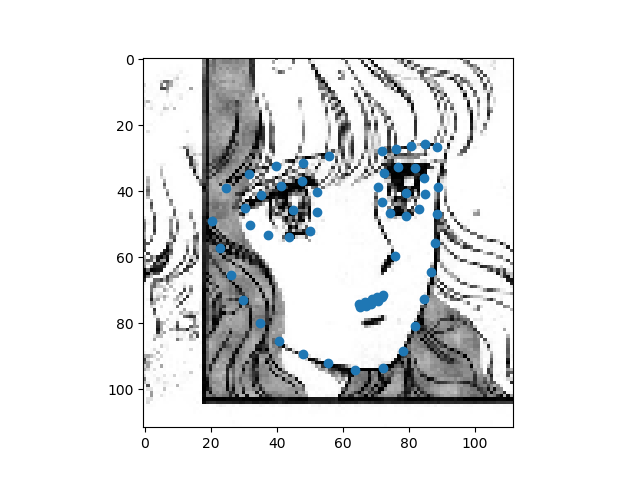}
\includegraphics[width=.4\textwidth]{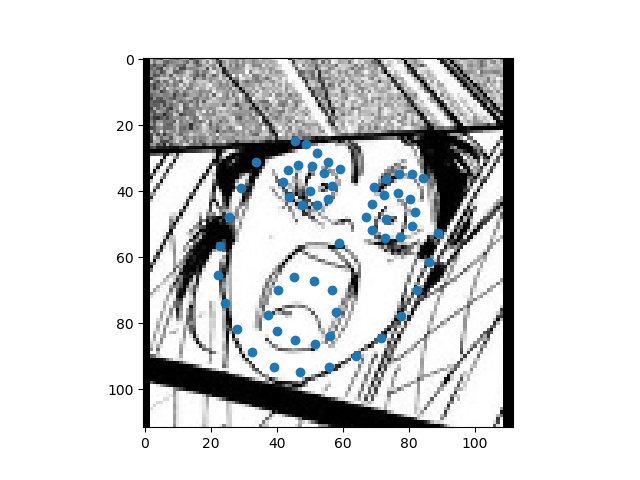}\\
\includegraphics[width=.4\textwidth]{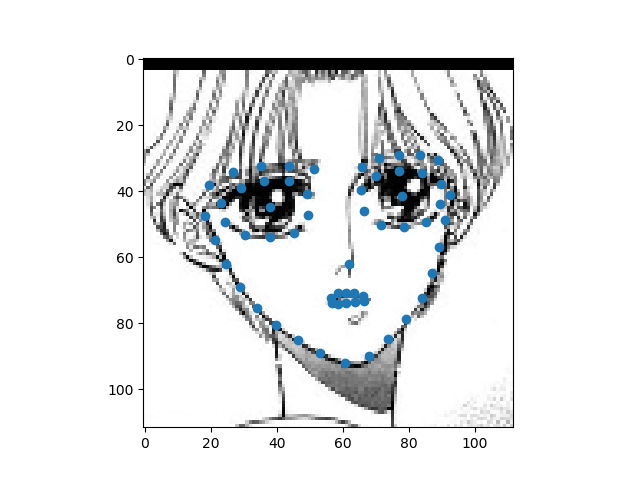}
\includegraphics[width=.4\textwidth]{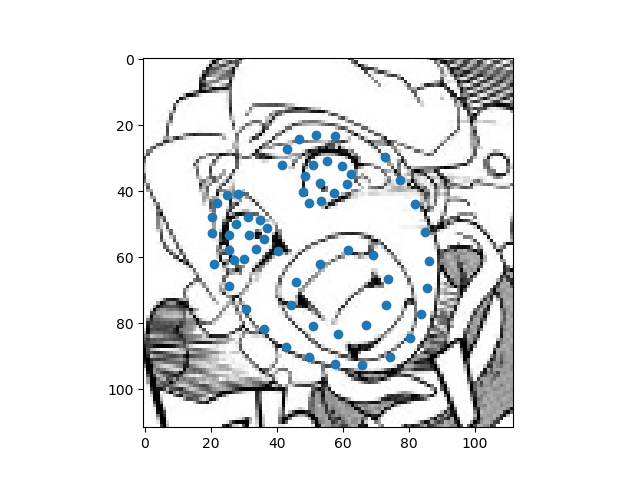}
\caption{Visualization of the best and worst landmark predictions. The left column shows the best predictions while the right column shows the worst predictions.
The images from the left, top to bottom are taken from: PrismHeart by Asatsuki Mai, 2nd and 3rd are Ningyoushi by Omi Ayuko, PlatinumJungle by Shinohara Masami.
The images from the right, top to bottom are taken from: RinToSiteShippuNoNaka by Ide Chikae, OL\_Lunch by Sanri Youko, RisingGirl by Hikochi Sakuya, ParaisoRoad by Kanno Hiroshi}
\label{fig:successFailure}
\end{figure}

Figure~\ref{fig:successFailure} is showing the best success and failure cases for the two-stage DAN using augmented data.
As we can see, our method performs well for cases where the face is in a normal frontal pose. However it is difficult to find the landmarks on drawings where the facial expressions are exaggerated such as a large and open mouth or closed eyes.

\section{Conclusions}\label{sec:conclusion}
This paper presents three contributions to the facial landmark detection for manga images: (1) we created a landmark annotation model inspired by iBUG for manga faces, (2) we created a manga faces landmark dataset and opened it to the public to encourage further research on this topic, and (3) we implemented one of the first landmark detection model for manga faces and show encouraging results. 

Our results can be used to further improve a landmark detection system for manga faces. Furthermore such landmark information can be used for many different applications, for example in improving the accuracy of character recognition systems as it has been shown in~\cite{jha2018towards} 
 or as further information to improve other applications such as emotion recognition. The landmarks can also be used for animating the face of the character or to transfer the emotion or style of one character to another one. 
\bibliographystyle{spmpsci}      
\bibliography{egbib.bib}

\begin{thebibliography}{10}
\providecommand{\url}[1]{{#1}}
\providecommand{\urlprefix}{URL }
\expandafter\ifx\csname urlstyle\endcsname\relax
  \providecommand{\doi}[1]{DOI~\discretionary{}{}{}#1}\else
  \providecommand{\doi}{DOI~\discretionary{}{}{}\begingroup
  \urlstyle{rm}\Url}\fi

\bibitem{amos2016openface}
Amos, B., Ludwiczuk, B., Satyanarayanan, M.: Openface: A general-purpose face
  recognition library with mobile applications  (2016)

\bibitem{augereau2018survey}
Augereau, O., Iwata, M., Kise, K.: A survey of comics research in computer
  science.
\newblock Journal of Imaging \textbf{4}(7) (2018)

\bibitem{augereau2016comic}
Augereau, O., Matsubara, M., Kise, K.: Comic visualization on smartphones based
  on eye tracking.
\newblock In: Proceedings of the 1st International Workshop on coMics ANalysis,
  Processing and Understanding, p.~4. ACM (2016)

\bibitem{bulat2017far}
Bulat, A., Tzimiropoulos, G.: How far are we from solving the 2d \& 3d face
  alignment problem?(and a dataset of 230,000 3d facial landmarks).
\newblock In: International Conference on Computer Vision, vol.~1, p.~8 (2017)

\bibitem{chu2017manga}
Chu, W.T., Li, W.W.: Manga facenet: Face detection in manga based on deep
  neural network.
\newblock In: Proceedings of the 2017 ACM on International Conference on
  Multimedia Retrieval, pp. 412--415. ACM (2017)

\bibitem{MangaFace}
Chu, W.T., Li, W.W.: Manga facenet: Face detection in manga based on deep
  neural network (2017).
\newblock \urlprefix\url{https://www.cs.ccu.edu.tw/~wtchu/projects/MangaFace/}

\bibitem{daiku2017comic}
Daiku, Y., Augereau, O., Iwata, M., Kise, K.: Comic story analysis based on
  genre classification.
\newblock In: Document Analysis and Recognition (ICDAR), 2017 14th IAPR
  International Conference on, vol.~3, pp. 60--65. IEEE (2017)

\bibitem{gupta2018imagine}
Gupta, T., Schwenk, D., Farhadi, A., Hoiem, D., Kembhavi, A.: Imagine this!
  scripts to compositions to videos.
\newblock arXiv preprint arXiv:1804.03608  (2018)

\bibitem{inoue2018cross}
Inoue, N., Furuta, R., Yamasaki, T., Aizawa, K.: Cross-domain weakly-supervised
  object detection through progressive domain adaptation.
\newblock arXiv preprint arXiv:1803.11365  (2018)

\bibitem{jha2018towards}
Jha, S., Agarwal, N., Agarwal, S.: Towards improved cartoon face detection and
  recognition systems.
\newblock arXiv preprint arXiv:1804.01753  (2018)

\bibitem{kasinski2008put}
Kasinski, A., Florek, A., Schmidt, A.: The put face database.
\newblock Image Processing and Communications \textbf{13}(3-4), 59--64 (2008)

\bibitem{kataoka2017automatic}
Kataoka, Y., Matsubara, T., Uehara, K.: Automatic manga colorization with color
  style by generative adversarial nets.
\newblock In: Software Engineering, Artificial Intelligence, Networking and
  Parallel/Distributed Computing (SNPD), 2017 18th IEEE/ACIS International
  Conference on, pp. 495--499. IEEE (2017)

\bibitem{kowalski2017deep}
Kowalski, M., Naruniec, J., Trzcinski, T.: Deep alignment network: A
  convolutional neural network for robust face alignment.
\newblock In: Proceedings of the International Conference on Computer Vision \&
  Pattern Recognition (CVPRW), Faces-in-the-wild Workshop/Challenge, vol.~3,
  p.~6 (2017)

\bibitem{le2012interactive}
Le, V., Brandt, J., Lin, Z., Bourdev, L., Huang, T.S.: Interactive facial
  feature localization.
\newblock In: European Conference on Computer Vision, pp. 679--692. Springer
  (2012)

\bibitem{lv2017deep}
Lv, J., Shao, X., Xing, J., Cheng, C., Zhou, X.: A deep regression architecture
  with two-stage re-initialization for high performance facial landmark
  detection.
\newblock In: Proceedings of IEEE Conference on Computer Vision and Pattern
  Recognition (2017)

\bibitem{matsui2017sketch}
Matsui, Y., Ito, K., Aramaki, Y., Fujimoto, A., Ogawa, T., Yamasaki, T.,
  Aizawa, K.: Sketch-based manga retrieval using manga109 dataset.
\newblock Multimedia Tools and Applications \textbf{76}(20), 21811--21838
  (2017)

\bibitem{rashid2017interspecies}
Rashid, M., Gu, X., Lee, Y.J.: Interspecies knowledge transfer for facial
  keypoint detection.
\newblock In: The IEEE Conference on Computer Vision and Pattern Recognition
  (CVPR), vol.~2 (2017)

\bibitem{ren2014face}
Ren, S., Cao, X., Wei, Y., Sun, J.: Face alignment at 3000 fps via regressing
  local binary features.
\newblock In: Proceedings of the IEEE Conference on Computer Vision and Pattern
  Recognition, pp. 1685--1692 (2014)

\bibitem{sagonas2016300}
Sagonas, C., Antonakos, E., Tzimiropoulos, G., Zafeiriou, S., Pantic, M.: 300
  faces in-the-wild challenge: Database and results.
\newblock Image and Vision Computing \textbf{47}, 3--18 (2016)

\bibitem{sagonas2013300}
Sagonas, C., Tzimiropoulos, G., Zafeiriou, S., Pantic, M.: 300 faces
  in-the-wild challenge: The first facial landmark localization challenge.
\newblock In: Computer Vision Workshops (ICCVW), 2013 IEEE International
  Conference on, pp. 397--403. IEEE (2013)

\bibitem{sagonas2013semi}
Sagonas, C., Tzimiropoulos, G., Zafeiriou, S., Pantic, M.: A semi-automatic
  methodology for facial landmark annotation.
\newblock In: Computer Vision and Pattern Recognition Workshops (CVPRW), 2013
  IEEE Conference on, pp. 896--903. IEEE (2013)

\bibitem{wu2017simultaneous}
Wu, Y., Gou, C., Ji, Q.: Simultaneous facial landmark detection, pose and
  deformation estimation under facial occlusion.
\newblock arXiv preprint arXiv:1709.08130  (2017)

\bibitem{xiong2013supervised}
Xiong, X., De~la Torre, F.: Supervised descent method and its applications to
  face alignment.
\newblock In: Computer Vision and Pattern Recognition (CVPR), 2013 IEEE
  Conference on, pp. 532--539. IEEE (2013)

\bibitem{yang2015facial}
Yang, J., Deng, J., Zhang, K., Liu, Q.: Facial shape tracking via
  spatio-temporal cascade shape regression.
\newblock In: Proceedings of the IEEE International Conference on Computer
  Vision Workshops, pp. 41--49 (2015)

\bibitem{zafeiriou2017menpo}
Zafeiriou, S., Trigeorgis, G., Chrysos, G., Deng, J., Shen, J.: The menpo
  facial landmark localisation challenge: A step towards the solution.
\newblock In: in Proc. IEEE Conf Comput. Vision Pattern Recognit. Workshops,
  pp. 2116--2125 (2017)

\bibitem{zhang2016joint}
Zhang, K., Zhang, Z., Li, Z., Qiao, Y.: Joint face detection and alignment
  using multitask cascaded convolutional networks.
\newblock IEEE Signal Processing Letters \textbf{23}(10), 1499--1503 (2016)

\end{thebibliography}

\end{document}